\documentclass[11pt]{article}
\usepackage{acl}
\usepackage{times}
\usepackage{latexsym}
\usepackage[T1]{fontenc}
\usepackage[utf8]{inputenc}
\usepackage{microtype}
\usepackage{inconsolata}
\usepackage{subfiles} 
\usepackage{amsmath}
\usepackage{amssymb}
\usepackage{amsthm}
\usepackage{enumerate}
\usepackage{graphicx}
\usepackage{multirow}
\usepackage{subfig}
\usepackage{pdfpages}
\usepackage{enumitem}
\usepackage{adjustbox}
\usepackage{authblk}
\usepackage{booktabs}
\usepackage{mathtools}
\def\method{FUSE}

\newtheorem{corollary}{Corollary}
\newtheorem{assumption}{Assumption}
\newtheorem{definition}{Definition}
\newtheorem{theorem}{Theorem}

\title{FUSE: Measure-Theoretic Compact Fuzzy Set Representation for Taxonomy Expansion}

\author{\textbf{Fred Xu}} 
\author{\textbf{Song Jiang}}
\author{\textbf{Zijie Huang}}
\author{\textbf{Xiao Luo}}
\author{\textbf{Shichang Zhang}}
\author{\textbf{Yuanzhou Chen}}
\author{\textbf{Yizhou Sun}}
\affil{Department of Computer Science, University of California, Los Angeles }
\affil{\{fredxu, songjiang, zijiehuang, xiaoluo, shichang, adrianchen, yzsun\}@cs.ucla.edu}
\begin{document}

\maketitle

\begin{abstract}
Taxonomy Expansion, which models complex concepts and their relations, can be formulated as a set representation learning task. The generalization of set, fuzzy set, incorporates uncertainty and measures the information within a semantic concept, making it suitable for concept modeling. Existing works usually model sets as vectors or geometric objects such as boxes, which are not closed under set operations. In this work, we propose a sound and efficient formulation of set representation learning based on its volume approximation as a fuzzy set. The resulting embedding framework, \textit{\underline{Fu}zzy \underline{S}et \underline{E}mbedding} (FUSE), satisfies all set operations and compactly approximates the underlying fuzzy set, hence preserving information while being efficient to learn, relying on minimum neural architecture. We empirically demonstrate the power of FUSE~on the task of taxonomy expansion, where FUSE~achieves remarkable improvements up to 23\% compared with existing baselines. Our work marks the first attempt to understand and efficiently compute the embeddings of fuzzy sets.
\end{abstract}

\section{Introduction}\label{sec:1}

Taxonomy is a crucial data structure for modeling semantic concepts, hence of great importance for NLP~\citep{lu2023facilitating,xu2023tacoprompt,yu2023fine}. Concepts in a taxonomy can often be viewed as sets, the most fundamental object in mathematics, whose operations directly link to First Order Logic (FOL). For example, in a science taxonomy, “Biology” and “Computer Science” are semantic concepts, whose intersection results in a new concept “Bio-informatics”, and “Diffusion Model” and “GAN” belong to a coarser-grained concept, “Generative Model”. Usually, sets are seen as a \textit{fixed collection} of objects. For example, the set $\mathbb{N}$ consists numbers $\{0, 1, \cdots\}$ by definition. However, in the context of semantic concepts, their meanings can change overtime and incorporate ambiguity. For example, ``beauty" is a concept that has become broader overtime, and ``deep learning models" can expand to have more elements with more discoveries made by the community. This underlying uncertainty and ambiguity are instead captured by a fuzzy set \citep{ZADEH1999, ZADEH19783}, an extension of classical sets. 

A wide range of works have been developed for set representation learning. Early efforts are made to construct simple vector embeddings \citep{mikolov2013efficient, pennington2014glove, devlin2019bert, vaswani2023attention, radford2018improving} based on similarity measures. To better model complex relationships such as asymmetrical relationships between concepts, geometric embeddings \citep{song-boxtaxo, hamilton2019embedding, ren2020query2box, ren2020beta} have been developed, which leverages the inherent geometric properties to model hierarchical relationships. However, these methods cannot address all the set operations including intersection, union, and complement. For example, box embedding \citep{song-boxtaxo, ren2020query2box, concept2box} doesn't define union and complement of boxes. Worse yet, existing geometric objects are not \textit{closed} under set operations: the union of two boxes is no necessarily a box, which can compromise the consistency of reasoning in the embedding space. 

In this paper, we directly tackle the challenge of fuzzy set representation learning for concept modeling. Our objective is to use their volume to quantify their information and their associated uncertainty. However, learning powerful representations for fuzzy sets is challenging. First, although extensive efforts have been made to incorporate deep learning techniques into fuzzy set modeling~\citep{chen2022fuzzy, dasgupta2022word2box,zhu2022neuralsymbolic}, their training procedure could be expensive when the universe of discourse is large. Second, compared with geometric embeddings, which have clear definitions of volume, it is unclear how to model the volumes of fuzzy sets due to the introduction of uncertainty and their abstract nature.

To tackle the previous challenges, we propose a principled and learnable model named \textit{\underline{Fu}zzy \underline{S}et \underline{E}mbedding (FUSE)} for fuzzy set representation learning. The core of \method{} is to introduce a compact approximation of fuzzy sets and then prove that \method{} can arbitrarily approximate the original fuzzy sets under reasonable regularity conditions. \method{} avoids the computational burden of accounting for all the elements in the space of discourse at once, while enjoying the properties of fuzzy logic, hence satisfying all set operations. We further introduce a rank-based loss and asymmetric relations to enhance set representation learning. To validate the effectiveness of our proposed \method{}, we evaluate on taxonomy expansion task and show that \method{} can achieve the performance improvement up to 23\% compared with state-of-the-art baselines, and we explore the effectiveness of our theoretical formulation through various ablation studies. 

Our \textbf{main contributions} can be summarized as follows: (a) We propose an embedding framework to model fuzzy sets and show that the embeddings satisfy all set operations and are closed under set operations. (b) We systematically construct this embedding as a proper approximation of fuzzy sets. (c) We demonstrate the effectiveness of this embedding framework on taxonomy expansion by comparing it against previous vector and geometry-based embedding methods.

\section{Related Work}
\subsection{Taxonomy Expansion}
 Taxonomy organizes concepts as a hierarchical graph, where nodes are concepts and edges denote “is-a” relationships between parent and child nodes.  As new knowledge is emerging, taxonomy expansion seeks to expand existing taxonomy with new nodes, which is a fundamental task for many real-world applications such as information filtering and recommendation. Existing works have focused on using a lexical vector representation in the spirit of language modeling and word embedding \citep{chang2018distributional,Snow-2004,mikolov2013efficient,pennington2014glove}. More recently, geometric embeddings such as box embedding has been used to better model the asymmetric relationship between parent and child nodes \citep{song-boxtaxo}. Compared to vector-based representations, they improved both the predictive performance and interpretability of the learned embeddings.\par  
 
\subsection{Set Representation Learning}
Set representation learning seeks to learn low-dimensional representations of data with a notion of volume and coverage. It is desirable when the representations can capture the rich semantic information and the complex relationships of data~\citep{rossi2020structural,wang2021structure,zhang2022benchmark,zhong2023knowledge}. For example, language modeling \citep{devlin2019bert, vaswani2023attention, radford2018improving} has aimed to learn vectors to represent combinatorically intractable combinations of human languages. In this context, semantic concepts can be viewed as sets. Recently, geometry-based approaches \citep{ren2020query2box, dasgupta2022word2box, ren2020beta,chen2021boxukg} have further improved the efficiency of the representations by enabling set operation such as intersection, but they fail to cover all operations and are not closed under them. Fuzzy set theory has explicitly formulated a way to represent the ambiguity of sets such as concepts in taxonomy construction, while automatically satisfying all desired properties of sets~\citep{chen2022fuzzy, zhu2022neuralsymbolic}. It is an extension to classical set theory with extensive applications~\citep{van2022analyzing,wagner2022neural,liang2023logic,yu2022probabilistic,xu2022neural}. 
For example, \citet{boratko-fuzzy} and \citet{dasgupta2022word2box} have explored the connection between fuzzy sets and box embeddings to model words.
However, existing fuzzy set representations lack a principled approach on what the low-dimensional representation stands for, and can be inefficient when the number of sets to model increases. We propose a novel solution by identifying the central characterization of a fuzzy set as its volume and approximate it using a compact representation, while yielding superior performance on the set representation learning task of taxonomy expansion.

\section{Preliminary}\label{sec:3}
\subsection{Fuzzy Sets} 
In contrast to classical set theory, which assigns a Boolean value to whether an element belongs to a set, a fuzzy set \citep{ZADEH19783} assigns a value between $0$ and $1$ to denote a \textit{degree of membership}. For a universe of discourse $U$, a fuzzy set is mathematically defined as a tuple $A = (U,m_A)$, where $A\subseteq U$ and $m_A: U \rightarrow [0,1]$ is its membership function. For any element $x\in U$, $m_A(x)$ represents the degree of membership of element $x$ in $A$. 
Fuzzy set models the uncertainty of membership by encoding imprecision and ambiguity in concepts. As an example, it can be used to describe the compatibility between two concepts, such as ``is-a" relationship. For example, for a concept ``Kobe Bryant" and a set of entities \{Basketball Player, Team Owner, Entrepreneur\}, a fuzzy membership function can be represented as the set $\{1.0, 0.1, 0.9\}$, each signifying ``Kobe Bryant"'s compatibility with each of the concepts in the set. 

Similar to standard sets, intersection, union, and complement between fuzzy sets are defined. Fuzzy set is related to fuzzy logic, which defines logical operations over soft truth values and follows G\"odel, product, or Łukasiewicz systems. For a detailed discussion of fuzzy logic systems, see \citet{chen2022fuzzy}. For language modeling, fuzzy sets can be used to model the ambiguity of the semantic meanings of words \citep{dasgupta-etal-2022-word2box}. In  taxonomy, fuzzy sets can be used to model concepts. 

\subsection{Possibility Theory} 
The membership function $m_A$ associated with a fuzzy set $A$ is constructed based on the theory of possibility in  \citet{ZADEH1999, ZADEH19783}. In the formulation of \citet{ZADEH19783}, to reason about linguistic concepts such as “likely”, a fuzzy set can be endowed with a probability-possibility distribution: 



\begin{definition}[\textbf{Possibility-Probability Distribution}]\label{def:pos-prob-dist}
    Let $U$ be the universe of discourse, and $(U,\mathcal{F},P)$ be a probability space, where $\mathcal{F}$ is the sigma-algebra and $P$ is the probability measure. 
    Let $X$ be a fuzzy variable that can take any values $x\in U$, and let $F$ be a fuzzy subset of $U$ with membership function $m_F$, then the \textbf{possibility of probability} of $X$ with respect to $F$ is:
    $$
    \pi_{P,X} = \int_U \pi_X dP = \int_U m_F dP.
    $$
    where $\pi_X$ is the possibility distribution associated with $X$ and $\int_U m_F dP$ is the Lebesgue integral of the membership function w.r.t to the probability measure $P$.
\end{definition}

Here the Lebesgue integral $\int_U m_F dP$,  in the sense of fuzzy set theory, represents the amount of information contained by the fuzzy set $F$. This construct can be seen as the measurement of information and uncertainty in the fuzzy variable $X$, making it a desirable quantity to approximate when learning a low-dimensional embedding of a fuzzy set. In our Fuzzy Set Embedding, we generalize definition \ref{def:pos-prob-dist} in definition \ref{def:fuzzy-measure}.

\section{Proposed Framework: FUSE}\label{sec:method}
\begin{figure*}
    \centering
    \includegraphics[width=0.8\textwidth]{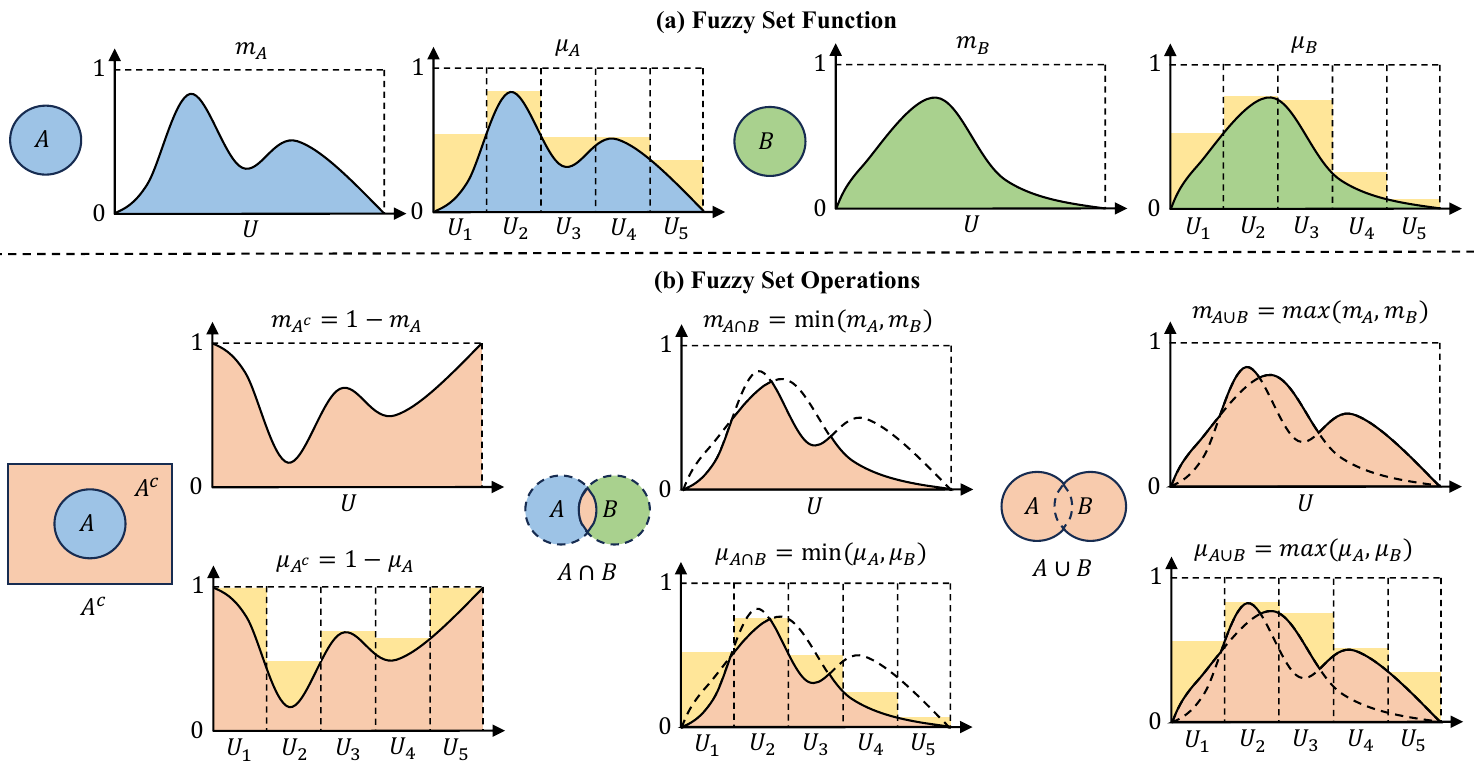}
    \caption{Illustration of set operations under fuzzy set membership function for two sets $A,B\in U$. $m_A$ is the fuzzy set membership function and $\mu_A$ the corresponding simple membership function (Definition \ref{def:sfs}). By using the fuzzy set representation, all the set operations (intersection, union, complement) are well-defined (G\"{o}del definition is used for easier illustration), and after set operations, results are still fuzzy set. For illustration, here we partition the universe $U$ into 5 partitions $\{U_1,\cdots, U_5\}.$}
    \label{fig:1}
\end{figure*}
We  now present \textbf{Fuzzy Set Embedding (FUSE)} for learning set representations in a principled way.


\subsection{Fuzzy Set Embedding}

To construct a proper embedding for fuzzy sets, we assume that the universe of discourse $U$ admits a finite partition, $\{U_i\}_{i=1}^d$, such that $U = \bigcup_{i=1}^d U_i$, and $U_i,U_j$ are disjoint if $i\neq j$. For a formal description of this assumption, see Appendix \ref{as}. In particular, this indicates that fuzzy set membership function $m_A$ has an associated \textit{simple function}:


\begin{definition}[\textbf{Simple Fuzzy Set}]\label{def:sfs}
Let $(U, \mathcal{F}, \xi)$ be a measure space and Let $U = \bigcup_{i=1}^d U_i$ be a finite partition of the universe $U$, and let $A \in \mathcal{F}$ and $m_A$ its membership function, then the \textbf{Simple Fuzzy Set} associated with $A$ is the tuple $(U, \mu_A)$, where:
\begin{equation}\label{eq:1}
\mu_A(x) := \sum_{i=1}^d \mathbf{1}_{\{x\in U_i\}} \mu_A^{(i)}(x), \forall x \in U
\end{equation}
is the \textbf{Simple Membership Function} of $A$, where $\mathbf{1}$ is the indicator function and $\forall x\in U, \forall i \in \{1,\cdots, d\}$:
\begin{equation}\label{eq:2}
\mu_A^{(i)}(x) := \sup_{x\in U_i} m_A(x). 
\end{equation}
\end{definition}

$\mu_A$ can be summarized in $d$ values $\mu_A^{(1)},\cdots, \mu_A^{(d)}$, each determined by the supremum of $m_A$ in the corresponding partition. To facilitate the standard application in deep learning, we formulate an alternative  representation in vector form to distinguish it from the functional representation denoted in Eqn. \ref{eq:1}.
\begin{definition}[\textbf{Fuzzy Set Embedding}] Let $A = (U,\mu_A)$ be a simple fuzzy set defined on the measure space $(U,\mathcal{F},\xi)$, where $U = \bigcup_{i=1}^d U_i$, then its corresponding \textbf{Fuzzy Set Embedding} (FUSE) is the $d$-dimensional vector: 
\begin{equation}\label{eq:fuse}
    \mathcal{U}_A := [\mu_A^{(1)},\cdots, \mu_A^{(d)}],
\end{equation}
\end{definition}

Since we are reducing the reasoning space from $[0,1]^{|U|}$ to $[0,1]^d$, we need to examine the loss incurred by this reduction. To reason about it in detail, we provide the following definition inspired by \citet{ZADEH19783, NAHMIAS197897} to quantify the amount of information covered by the fuzzy sets across the entire universe $U$. 

\begin{definition}[\textbf{Simple Fuzzy Measure}]\label{def:fuzzy-measure}
    Let $U$ be a compact universe of discourse and let $A=(U, m_A)$ be a fuzzy subset of $U$. Let $(U,\mathcal{F}, \xi)$ be a measure space defined on $U$, then the \textbf{fuzzy measure} of the fuzzy set $(U,m_A)$ is: 
    $$\mathbb{P}(A) :=  \int_U m_A d\xi.$$ 
    Then a \textbf{Simple Fuzzy Measure} of a simple fuzzy set $A=(U,\mu_A)$ is defined as: 
    $$
    \mathbb{P}_\mu (A) := \int_U \mu_A d\xi. 
    $$
\end{definition}
Given a finite partition, furthermore:
\begin{align}\label{eq:3}
    \mathbb{P}_\mu (A) = \sum_{i=1}^d \int_{U_i} \mu_A^{(i)} d\xi  = \sum_{i=1}^d \mu_A^{(i)} \xi(U_i),
\end{align}
where $\xi(U_i)$ corresponds to the measure of partition set $U_i$. If $\xi$ is a probability measure, then Definition \ref{def:fuzzy-measure} corresponds to Definition \ref{def:pos-prob-dist}. In practice, we examine choices of different measures empirically in Section \ref{sec:5}. In short, a simple fuzzy set $A=(U,\mu_A)$ approximates the fuzzy measure of the underlying fuzzy set $(U,m_A)$. We state this observation formally in theorem \ref{thm:monotone-convergence} and illustrates it in Figure \ref{fig:2}(a).

\begin{theorem}\label{thm:monotone-convergence}
    Let $U$ be a compact universe of discourse and $(U,\mathcal{F},\xi)$ a measure space. Let $A$ be a fuzzy subset of $U$ and $m_A$ its membership function that's measurable. Moreover, let $\mu_A$ be its simple membership function, then $\forall \epsilon > 0$, $\exists \delta > 0, d > 0$ such that if $d\delta = \xi(U)$ and $||U_i|| := \min_{i} \xi(U_i) < \delta$, we have: 
    $$
    0 < \mathbb{P}_\mu(A) - \mathbb{P}(A) < \epsilon.
    $$
\end{theorem}

The fuzzy measure of a simple fuzzy set is an upperbound for the fuzzy measure of its underlying fuzzy set and converges to the possibility of its underlying fuzzy set when the partition is sufficiently fine-grained. With suitable assumption on the function $m_A$, we can further establish the rate of convergence:


\begin{figure}
    \centering \includegraphics[width=0.5\textwidth]{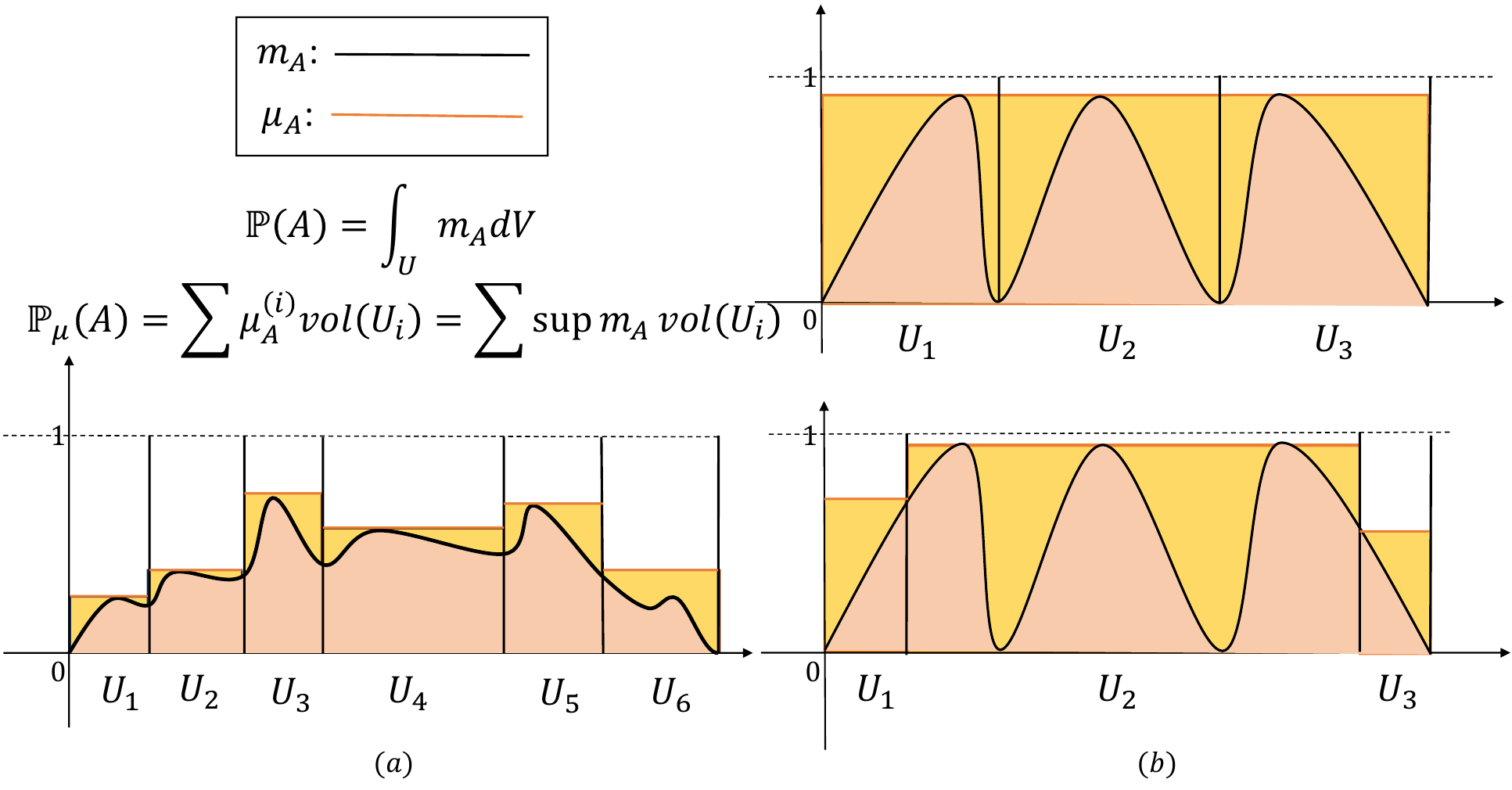}
    \caption{On the left plot (a), we illustrate when the universe is a compact subset of $\mathbb{R}$. Fuzzy measure $\mathbb{P}$ is the Riemann Integral of $m_A$ on embedding space $\mathcal{U}$ (the orange region), 
    while the Simple Fuzzy Measure $\mathbb{P}_\mu$ is its upper Darboux sum (the yellow region). On right plot (b), we demonstrate that for a fixed number of partitions, different choices of partition size result in different approximation of $\mathbb{P}$: the bottom-right partition has better approximation than the top-right partition, since it results in less over-estimation.}
    \label{fig:2}
\end{figure}

\begin{corollary} \label{cor:4.8}
    Following the condition in definition \ref{def:fuzzy-measure}, if in addition $m_A$ is Lipschitz-continuous or of bounded variation, then the convergence rate in \ref{thm:monotone-convergence} is $O(1/n)$, where $n$ is the number of partitions.
\end{corollary}

\subsection{Embedding-based Fuzzy Set Operators} 
\textbf{FUSE} combines set theory and measure theory and provides a theoretically sound embedding. For an entity $x \in U$, we treat it as a concept and associate with it a fuzzy subset $(U, m_A)$, representing its compatibility with other concepts. We can treat every entity as a fuzzy set embedding, define set operations in the language of set theory, and compute them using vector operations. Suppose we have two entities $x,y \in U$, and  $\mathcal{U}_A, \mathcal{U}_B \in [0,1]^d$ are the two fuzzy set embeddings associated with them, we can define following operations: 
\begin{itemize}[leftmargin=*]
    \item \textbf{Fuzzy Mapping}: Every entity/element is a singleton set, and $\mathcal{M}$ maps an entity $x \in U$ to its associated fuzzy set embedding $\mathcal{U}_{\{x\}} \in [0,1]^d$. In the case of taxonomy expansion, the input is the word vector $\mathbf{x}\in \mathbb{R}^e$ obtained from a pre-trained language model like BERT \citep{devlin2019bert}. To construct a map between the word vector and its  associated fuzzy set embedding, we use a neural networks $f: \mathbb{R}^e \rightarrow [0,1]^d$: 
\begin{align}
    \mathcal{U}_{A}=\mathcal{M}(A;\theta) = \sigma(f(\mathbf{x}; \theta)) \in [0,1]^d, 
\end{align}
where $\sigma$ is a normalization constraint to make the embedding space compact, such as sigmoid, 0-1 clamping, or Layernorm \citep{ba2016layer}. 
\end{itemize}
We can futher define set operations by product logic. Other fuzzy systems, such as G\"{o}del logic, is illustrated in Fig. \ref{fig:1}. 
\begin{itemize}[leftmargin=*]
    \item \textbf{Intersection}: 
    The intersection between the two fuzzy sets $A\cap B$ can be computed by element-wise product t-norm \citep{klement2013triangular}: 
    \begin{align}\label{eq:4}
        \mathcal{U}_{A\cap B} = 
        \mathcal{U}_A \odot \mathcal{U}_B.
    \end{align}
    where $\odot$ is element-wise multiplication.
    \item \textbf{Union}: The union of two fuzzy sets $A\cup B$ can be computed by element-wise product t-conorm:
    \begin{align}\label{eq:5}
         \mathcal{U}_{A\cup B} = \mathcal{U}_A + \mathcal{U}_B - \mathcal{U}_A \odot \mathcal{U}_B.
    \end{align}
    \item \textbf{Complement}: The complement of a fuzzy set $A$ denoted as $A^c$ can be computed as: 
    \begin{align}\label{eq:6}
        \mathcal{U}_{A^c}= \mathbf{1} - \mathcal{U}_{A}.
    \end{align}
\end{itemize}



\begin{figure*}
    \centering
    \includegraphics[width=0.75\textwidth]{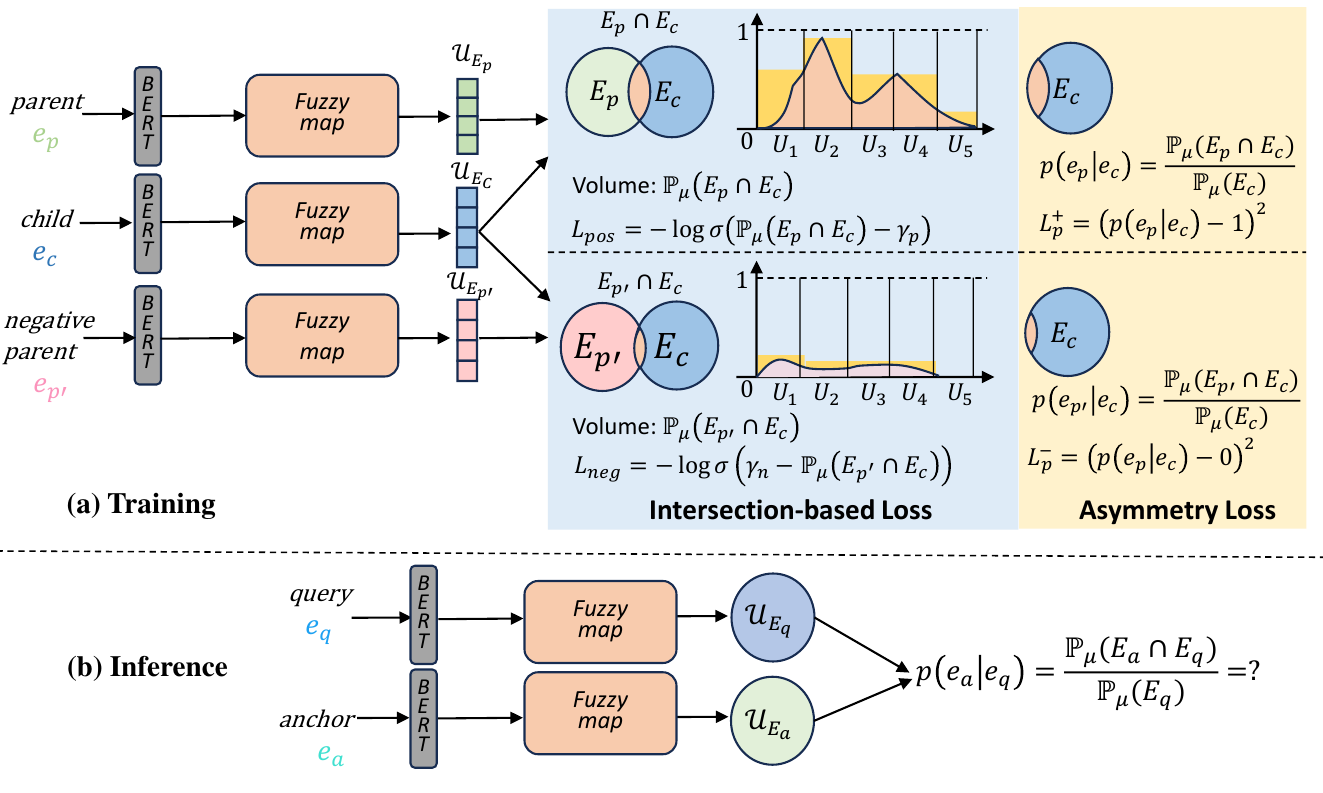}
    \caption{The overview of FUSE used to model concepts in taxonomy. First the entities are converted into vectors using Bert, then the main Fuzzy Map transforms the vector into a fuzzy set embedding. (a) Training: FUSE is learned using both a volume-based intersection loss and a volume-based asymmetry loss. (b) Inference: use the taxonomy probability score to check the degree of containment of a query in the anchor. Here $e$ are entities, $E$ are their associated fuzzy sets, and $\mathcal{U}$ the fuzzy set embeddings.}
    \label{fig:3}\vspace{-10pt}
\end{figure*}

\subsection{Taxonomy Expansion with FUSE}
In this part, we use \emph{taxonomy expansion} task to showcase the advantages of representing concepts with fuzzy set embeddings. 

\textbf{Membership Prediction with FUSE.} After representing a concept with fuzzy sets, the core task of taxonomy expansion is to determine whether an element $y$ belongs to a set $A$, and this is often used as a score function in pair-based relationship in taxonomy expansion task \citep{song-boxtaxo, Shen_2020, Yu_2020}. Using our framework, for some element $y\in U$, we can apply the entity mapping function $\mathcal{M}$ to find its fuzzy set embedding $\mathcal{U}_{\{y\}}$. Then we can simply measure the degree of membership of element $y$ in some other fuzzy set $A$ by considering the fuzzy measure of the fuzzy set embedding $\mathcal{U}_{A\cap \{y\}} = \mathcal{U}_A \odot \mathcal{U}_{\{y\}}$, which we denote as $\mathbb{P}_\mu(A\cap \{y\})$ and compute it as:
\begin{align}\label{eq:9}
 \mathbb{P}_{\mu}(A\cap \{y\}) = \sum_{i=1}^d \left(\mathcal{U}_A^{(i)} \mathcal{U}_{\{y\}}^{(i)} \right) \xi(U_i).
\end{align}
In training, we model the volume using global trainable weights $\textbf{w} = \{ w_1, \cdots, w_d\}$ with a normalization transform to restrict the type of measure $\xi$. Therefore, we approximate $\mathbb{P}_{A\cap \{y\}}$ and define the standard score function:
\begin{align}\label{eq:10}
     \psi(y, A) = &\sum_{i=1}^d \mu_{A\cap \{y\}}^{(i)} w_i  = (\mathcal{U}_A \odot \mathcal{U}_{\{y\}})^T \mathbf{w}.
\end{align}

and the corresponding ranking-based loss:
\begin{equation} \label{eq:12}
\begin{aligned}
    L(y,A) = &-\log \sigma\left(\psi(y,A)-\gamma_p\right) \\
     &-\frac{1}{k} \sum_{i=1}^k \log \sigma\left(\gamma_n - \psi(y',A)\right)
\end{aligned}
\end{equation}
where $(y,A)$ are positive pairs and $(y',A)$ negative pairs, and $\gamma_p, \gamma_n$ are margins for positive and negative predictions. We use different margins since by Theorem \ref{thm:monotone-convergence}, fuzzy measure of the fuzzy set embedding is an upperbound for the underlying fuzzy set, so the result we obtain is overestimating the actual fuzzy measure. We provide an ablation study regarding choice of margin in section \ref{sec:5}.

\textbf{Incorporating Asymmetric Relation.} As signified in \citet{song-boxtaxo}, membership prediction in a taxonomy expansion task usually involves asymmetric relations. For example, parent nodes in a taxonomy usually strictly incorporate the concept in the child nodes. Since the score function, which uses the intersection between two sets, is symmetric, here we propose another score function to signify this asymmetry. Suppose we have two entities  $e_p, e_c \subseteq U$, such that $e_c$ is a child of $e_p$. Let $E_p, E_c$ be their associated fuzzy sets, then we can model this relationship by:
\begin{align}\label{eq:13}
    P(e_p|e_c) = \frac{\mathbb{P}_\mu (E_p\cap E_c)}{\mathbb{P}_\mu (E_c)} = \frac{(\mathcal{U}_{E_p}\odot \mathcal{U}_{E_c})^T \mathbf{w}}{\mathcal{U}_{E_c}},
\end{align}
where we use the simple fuzzy measure for each set as its volume and use the ratio between the volume of $E_p\cap E_c$ and the volume of $E_c$ as the result. For a positive child-parent pair $(e_c,e_p)$, the loss is:
\begin{align}
    L^+_p = (P(e_p|e_c) - 1)^2,
\end{align}
whereas for a negative child-parent pair $(e_c,e_p')$:
\begin{align}
    L^-_p = (P(e_p'|e_c) - 0)^2. 
\end{align}
The main difference between Eqn. \ref{eq:10} and the case of box embedding in \citet{song-boxtaxo} is that the volume of a fuzzy set embedding spans the entire universe $U$. We combine the pair-based ranking loss and the asymmetric child-parent pair loss:
\begin{equation}\label{eq:14}
    L_{taxo} = L(e_c,e_p) + \lambda (L_p^+ +  L_p^-),
\end{equation}
where $\lambda$ is a hyper-parameter to control the strength of each loss. 

\begin{table}[h]
    \begin{adjustbox}{max width= 0.5\textwidth}
    \begin{tabular}{ccccccc}
    \toprule
     Dataset & \multicolumn{3}{c}{Environment} & \multicolumn{3}{c}{Science} \\
     \hline
     Metric & ACC & MRR & Wu\&P & ACC & MRR & Wu\&P \\\midrule
     TAXI & 16.7 & N/A & 44.7 & 13.0 & N/A & 32.9 \\
     HypeNet & 16.7 & 23.7 & 55.8 & 15.4 & 22.6 & 50.7 \\
     BERT+MLP & 11.1 & 21.5 & 47.9 & 11.5 & 15.7 & 43.6 \\ 
     TaxoExpan & 11.1 & 32.3 & 54.8 & 27.8 & 44.8 & 57.6 \\  
     STEAM & 36.1 & 46.9 & 69.6 & \underline{36.5} & \underline{48.3} & \underline{68.2} \\ 
     BoxTaxo & \underline{38.1} & \underline{47.1} & \underline{75.4} & 31.8 & 45.3 & 64.7\\\midrule
     FUSE & \textbf{42.3} & \textbf{58.3} & \textbf{77.6} & \textbf{39.9} & \textbf{52.9} & \textbf{73.4} \\
     FUSE ($\lambda=1.0$) & \textbf{43.1} & \textbf{53.3} &  74.3 &  \textbf{43.5} & \textbf{56.6} & \textbf{77.5}
    \\
    \bottomrule
    \end{tabular}
\end{adjustbox}
    \caption{Results on taxonomy expansion compared to existing methods. Here bold font refers to the best performance results (compared to baseline) while underline refers to the second-best performance result. The results are reported as average over 5 runs. ``N/A" is present since MRR is not applicable to TAXI. FUSE is our base model while FUSE ($\lambda =1.0$) is the model with balanced weights for intersection and asymmetric loss.}
    \label{table:taxo}
\end{table}

\section{Experiments}\label{sec:5}

\textbf{Dataset}: We use two public datasets (Environment, Science) from SemEval-16 taxonomy construction tasks. Following the training setup in \citet{song-boxtaxo}, we sample 20\% of the leaf nodes as test set and use the rest as training data. The performance of taxonomy expansion task can be found in table \ref{table:taxo}, where the results are averaged over 5 runs to reduce variance. 

\textbf{Metrics}: We use three metrics, Accuracy (ACC), Mean Reciprocal Rank (MRR), and Wu \& Palmer similarity (Wu\&P) to measure the performance for the baseline models and FUSE.  

\textbf{Baselines}: The baseline comparison models are vector-based models like TAXI \citep{panchenko-etal-2016-taxi}, HypeNet \citep{shwartz-etal-2016-improving}, BERT+MLP \citep{Yu_2020}, TaxoExpan\citep{Shen_2020}, STEAM \citep{Yu_2020}, and geometry-based model like BoxTaxo \citep{song-boxtaxo}.

\begin{figure}[h]
    \includegraphics[width=0.48\textwidth]{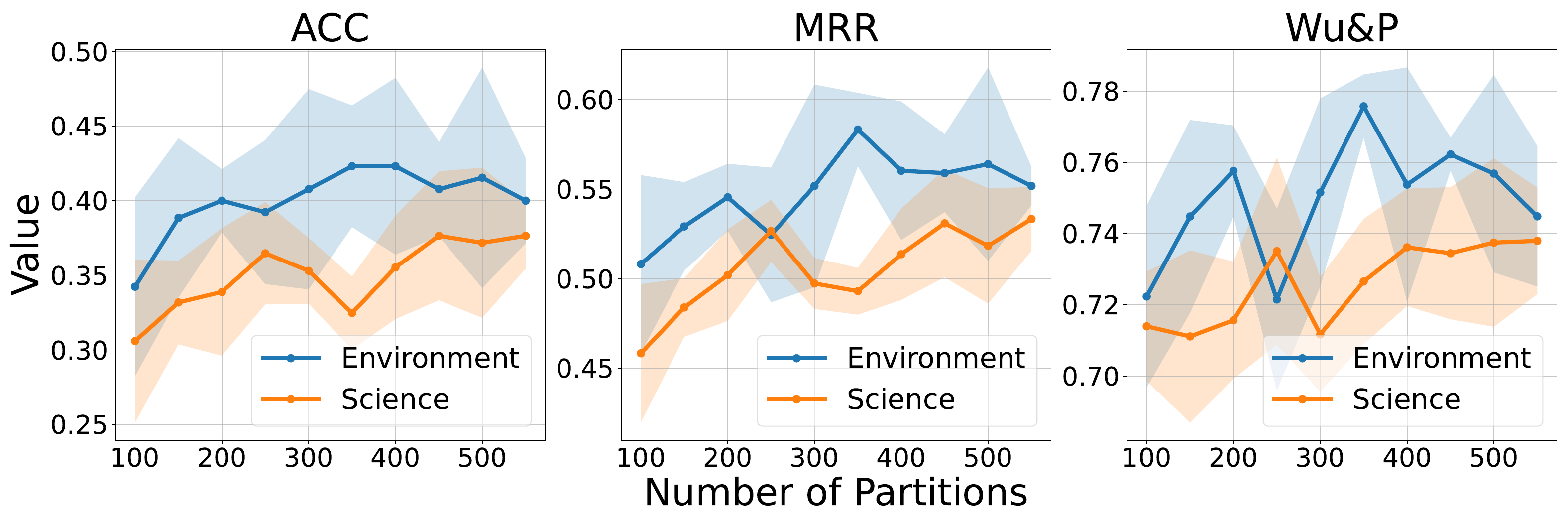}
    \vspace{-15pt}
    \caption{Trend of model performance with varying number of partitions on the science dataset.}
    \label{fig:partition_science}\vspace{-15pt}
\end{figure}
\subsection{Does FUSE Benefit from More Partitions?}

Here we examine the impact of choice of number of partitions, since according to theorem \ref{thm:monotone-convergence}, as $d$ increases, we should expect better approximation to the underlying fuzzy set. In this experiment, we vary the number of partitions from $100$ to $550$, with increment of $50$, and measure the performance of taxonomy expansion on both Science and Environment datasets, averaged over 5 runs. The resulting trend on both datasets can be seen in Figure \ref{fig:partition_science}.


In both cases, we can see a general upward trend on model's average performance when the number of partitions goes up. This provides empirical support for the theoretical result in theorem \ref{thm:monotone-convergence}. The variance of model performance tends to increase when the number of partitions goes up, which suggests that smaller learning rates and other normalization for optimization may be considered. 
 
\begin{table}[h]
        \begin{adjustbox}{max width= 0.5\textwidth}
    \begin{tabular}{ccccccc}
    \toprule
     Dataset & \multicolumn{3}{c}{Environment} & \multicolumn{3}{c}{Science} \\
     \hline
     Metric & ACC & MRR & Wu\&P & ACC & MRR & Wu\&P \\\midrule
     FUSE-sigmoid & \textbf{45.0} & \underline{57.2} & \underline{75.6} & \textbf{40.0}  & \underline{52.7} & \textbf{74.7} \\
     FUSE-softmax& 7.3 & 17.3 & 51.9 & 21.3 & 32.0 & 65.2 \\
     FUSE-01 & 41.3 & 55.8 & 75.4 & 37.3  & 52.3 & 72.3  \\ 
     FUSE & \underline{42.3} & \textbf{58.3} & \textbf{77.6} & \underline{39.9} & \textbf{52.9} & \underline{73.4} 
    \\
    \bottomrule
    \end{tabular}
    \vspace{-10pt}
    \end{adjustbox}
    \caption{Results on taxonomy expansion compared under different choices of normalization on volume weights. This corresponds to different choices of measure space.}
    \label{table:taxo-2}
\end{table}

\subsection{Does the Choice of Measure Affect Model Performance?}

As for the proposed new score function based on the volume of the fuzzy set in Eqn. \ref{eq:10}, we study the impact of different choices of measure $\xi$. This corresponds to different choices of normalization applied to the volume of each partition. If we follow definition \ref{def:pos-prob-dist}, then partition volumes follows a probability distribution, indicating a softmax normalization on the global weights. Otherwise, we can choose to use sigmoid or 0-1 clamping. Results over 5 runs for different choices of measures can be found in table \ref{table:taxo-2}. We observe that the softmax normalization, which enforces the volume weights to follow a probability distribution, doesn't work well overall. This suggest that the construction we proposed in definition \ref{def:fuzzy-measure} is more suitable than the classical definition in \ref{def:pos-prob-dist}. We also observe that using a sigmoid normalization on the volume weights can improve results. 

\subsection{Does Asymmetry Loss Help Taxonomy Expansion Task?}
\begin{figure}[h]
    \includegraphics[width=0.48\textwidth]{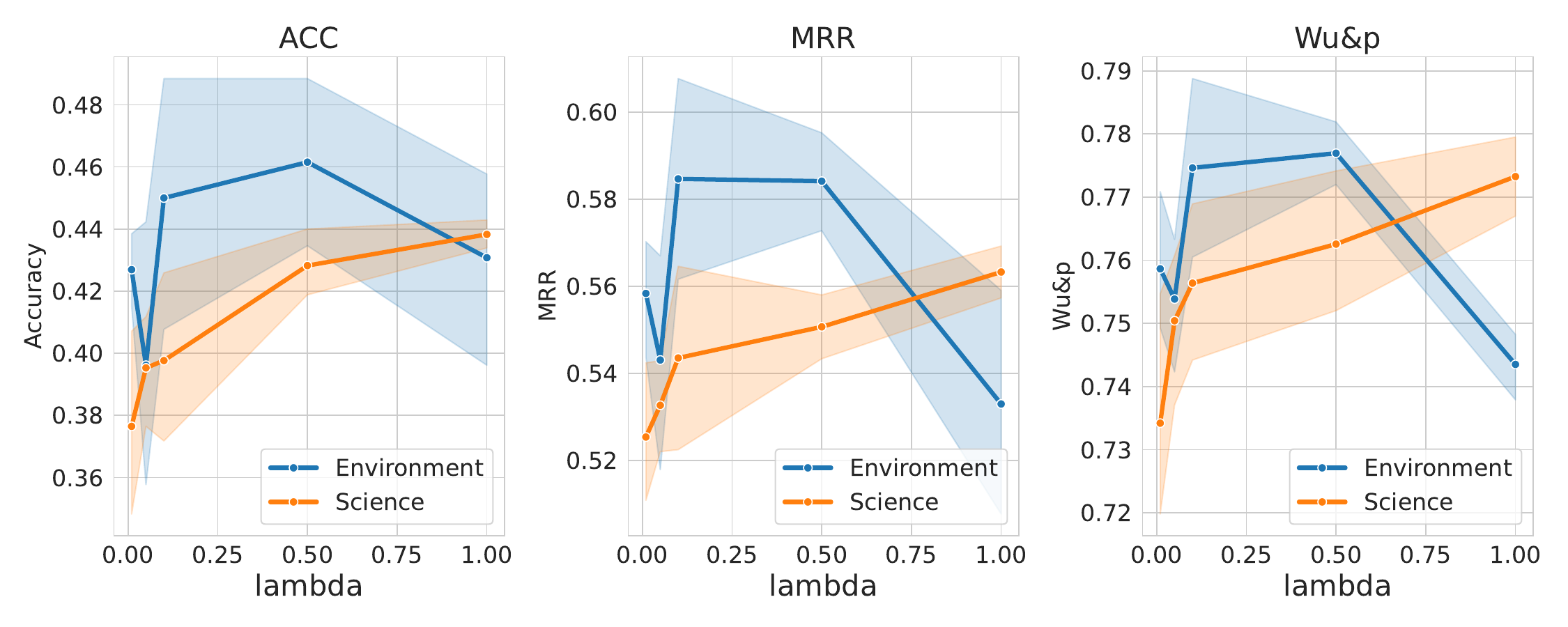}
    \vspace{-15pt}
    \caption{Model performance with different strength of asymmetry, lambda.}
    \label{fig:lambda}\vspace{-15pt}
\end{figure}

In this ablation study, we examine the impact of applying asymmetry losses, since set intersection is a symmetric operation, whereas the membership relation is asymmetric. The value of $\lambda$ in Eqn. \ref{eq:13} controls how much should the asymmetry-based loss affect the training (the greater the value of $\lambda$, the greater the impact). The results, averaged over 5 runs is in figure \ref{fig:lambda}. We can see a trend of performance improvement on Science dataset and an increase in performance for $\lambda > 0$ for the Environment dataset. This result validates the importance of modeling asymmetric relations.

\subsection{Does Wider Margin Affect Model Performance?}
\begin{figure}[h]
    \includegraphics[width=0.48\textwidth]{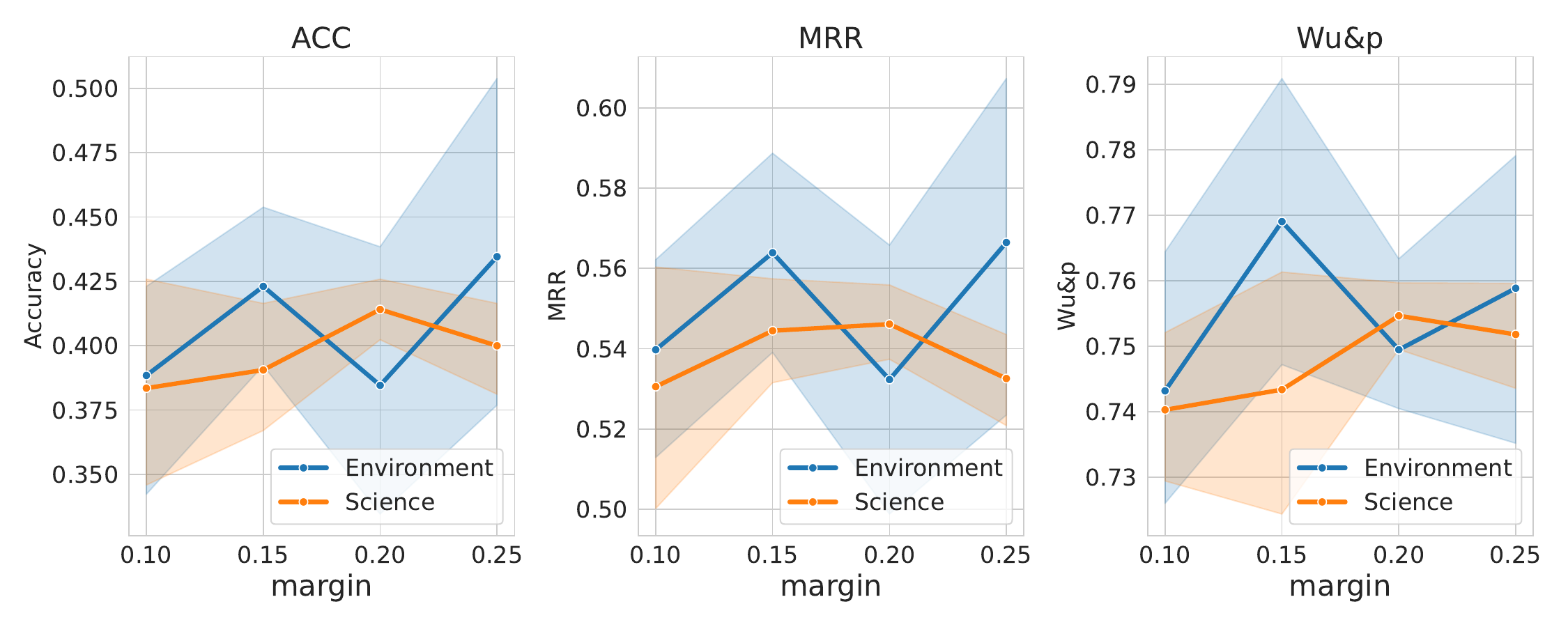}
    \vspace{-15pt}
    \caption{Model performance with different choice of margins.}
    \label{fig:margin}\vspace{-15pt}
\end{figure}

In this ablation study, we examine the impact of different margins on the learning performance, as presented in Eqn. \ref{eq:12}, since theoretically, our construction of FUSE over-estimates the volume under the fuzzy set. In figure \ref{fig:margin}, which presents the average result over 5 runs. From the result, we can see that having wider margin (in this case $\gamma_p = 0.6, \gamma_n = 0.4$) does benefit the model performance, supporting our hypothesis that FUSE over-estimates the volume of a fuzzy set. 
    




\subsection{Additional Experiment: Is there Synergy Between Different Modeling Choices?}
In this part, we examine the synergy between configurations across all the ablation studies performed. In this case we study the case where all the best configurations ($\lambda = 0.5, \Delta \gamma = 0.2$, sigmoid-normalization) are used. We call the resulting model FUSE-comb(ine) and report the performance averaged over 5 runs compared to other models in table \ref{table:taxo-5}. The result doesn't immediate suggest that combining multiple best case scenarios would result in optimal performance. The strongest performance so far comes from setting $\lambda=0.5$. 

\begin{table}[h]
    \centering
    \caption{Results on FUSE-comb compared with each individual best configurations}
    \begin{adjustbox}{max width= 0.5\textwidth}
    \begin{tabular}{ccccccc}
    \toprule
     Dataset & \multicolumn{3}{c}{Environment} & \multicolumn{3}{c}{Science} \\
     \hline
     Metric & ACC & MRR & Wu\&P & ACC & MRR & Wu\&P \\\midrule
     FUSE-sigmoid & \underline{45.0} & \underline{57.2} & 75.6 & 40.0  & 52.7 & 74.7 \\
     FUSE ($\lambda=0.5$) & \textbf{46.2} & \textbf{58.4} & \textbf{77.7} & \textbf{42.8} & \textbf{55.1} & \underline{76.3} \\ 
     FUSE ($\Delta \gamma = 0.2$) & 38.5 & 53.2 & 74.9  & 41.4 & \underline{54.6} & 75.5 \\ 
     FUSE-comb & 41.1 & 53.0 & \underline{76.4} & \underline{42.4} & 54.3 & \textbf{76.7} \\ 
    \bottomrule
    \end{tabular}
    \end{adjustbox}
    \label{table:taxo-5}
\end{table}

\subsection{Infer about Union and Complement Using Trained Embeddings}

 In our taxonomy expansion experiment, the model is trained using only the volume-based intersection and asymmetry losses, without observing pairs of sets that are related by union or complements. In this case study we examine whether our embedding can generalize to these two operations. 

\textbf{Infer about Set Union}: For a parent entity $e_p$ in the taxonomy and its $m$ child entities $e_{c_1},\cdots, e_{c_m}$, we examine the similarity of union of fuzzy set embeddings of child entities with the fuzzy set embedding of the parent entity. That is, between $\mathcal{U}_{\bigcup_{i=1}^m E_{c_i}}$ and $\mathcal{U}_{E_p}$. To this end, we use the trained fuzzy set embedding from the FUSE ($\lambda = 1$) model and apply union operation in Eqn. \ref{eq:5} among all the child fuzzy set embeddings, then we rank the Euclidean distance between the obtained fuzzy set embedding (union of all child embeddings) against all the existing parent embeddings in the dataset. From \ref{table:taxo-5.6}, we observe that fuzzy set embedding captures union patterns. As an example from the Science dataset, for child entities [“calculus of variations", “analysis", “integral calculus"], with parent entity “calculus", the top-3 closest simple fuzzy set embedding corresponds to entity “calculus", “analysis", “geophysics", and the model's prediction is closest to the correct parent.

\textbf{Infer about Set Complement}: In this case we examine complement by the set operation $A\setminus B = A\cap B^c$. In particular, the fuzzy set for parent (denote it $A$) minus a child fuzzy set (denote it $B$) should be similar to the union of the remaining children fuzzy set. We follow the union, complement, and intersection operation to compute the fuzzy set embedding in this case, and again we use the embedding from FUSE ($\lambda = 1$) model. We compute the Euclidean distance between $A \cap B^c$ and all the existing child embeddings in the dataset. Here we present MRR and accuracy result in table \ref{table:taxo-5.6}. In contrast to union, it seems that complement doesn't achieve a reasonable performance. This may be due to the fact that the complement of a fuzzy set is taken over the entire universe of discourse, rather than simply in the scope of all the children entities.

\begin{table}[h]

    \begin{adjustbox}{max width=0.5\textwidth}
    \begin{tabular}{ccccc}
    
    \toprule
     Dataset & \multicolumn{2}{c}{Environment} & \multicolumn{2}{c}{Science} \\
     \hline
     Metric & ACC & MRR & ACC & MRR \\\midrule
     Union Inference & 81.3 & 85.3 & 85.4 & 89.5 \\
     Complement Inference &  2.9 & 14.8 & 11.9 & 26.6 \\
    \bottomrule
    \end{tabular}
    \end{adjustbox}
    \caption{Results on Union and Complement Inference using FUSE trained only with intersection based loss}
    \label{table:taxo-5.6}
\end{table}
 \vspace*{-\baselineskip}

\section{Conclusion}\label{sec:conclusion}
For taxonomy expansion, We propose a novel and theoretically sound Fuzzy Set Embedding (FUSE) to model concepts and relationship between concepts that incorporate set operations (intersection, union, complement). We show theoretically that FUSE preserves the information of the fuzzy set with sufficiently fine-grained partitions and demonstrate empirically that it can outperform existing vector-based and geometry-based embedding methods on taxonomy expansion. For future works, we believe that expanding the taxonomy dataset with more complicated combination of set operations, such as First Order Logic (FOL), can further improve the model performance.

\section{Limitations}
This work is the first attempt to use fuzzy set to model concepts in taxonomy expansion. To examine only the effectiveness of fuzzy set representation, we only use simple neural architectures and use only the child-parent pairs.  The full capacity of FUSE should be further examined using datasets that contain First Order Logic (FOL) statements, since by construction, fuzzy sets should satisfy all the fuzzy logic axioms \citep{chen2022fuzzy}. This suggests future directions to expand taxonomy datasets with more complicated queries, and to handle more graph-structured data in social analysis and text mining \citep{ren2020beta, ren2020query2box,chen2022fuzzy,zhu2022neuralsymbolic,ju2023comprehensive}. Moreover, we can explore more explicit form of fuzzy membership function, such as a mixture of Gumbel boxes, to make the learning more concrete. 

\section{Acknowledgments}

The authors thank all the  reviewers for
their valuable comments and constructive feedback.
The authors also acknowledge financial support from NSF 2211557, NSF 1937599, NSF 2119643, NSF 2303037, NSF 2312501, NASA, SRC JUMP 2.0 Center, Amazon Research Awards, and Snapchat Gifts.

\bibliography{anthology,custom}

\appendix

\section{List of Symbols and Notations}

In table \ref{table:a}, we list all the symbols and notations used in the paper, then we provide the theoretical proofs for the main results in the paper. 
\begin{table}
\centering
\begin{tabular}{@{}ll@{}} 
    \toprule
\textbf{Symbol} & \textbf{Description} \\
\midrule
$A,B,C,\cdots$ & mathematical sets.\\
$e_c, e_p, \cdots$ & entities in a taxonomy \\ 
$E_c, E_p, \cdots$ & fuzzy sets associated with entities \\ 
$U$ & The universe of discourse, \\
 & the set of all concepts. \\
$\mathbb{N}^+$ & The set of all positive integers. \\ 
$(U,\mathcal{F},\xi)$ & A measure space with universe $U$,\\
& sigma-algebra $\mathcal{F}$ and a measure $\xi.$  \\
$A^c$ & The complement of a set. \\
$A\cap B$ & Intersection of two sets.\\
$A\cup B$ & Union of two sets. \\
$(A,m_A)$ & a fuzzy set, where $A\subset U$, \\
& $m_A: U\rightarrow [0,1].$\\ 
$m_A$ & The fuzzy membership function 
 \\ 
 & associated with a fuzzy set. \\ 
$(A,\mu_A)$ & a simple fuzzy set, where $A\subset U$, \\ 
& $\mu_A: U\rightarrow [0,1]$.\\ 
$\mu_A$ & The simple membership function \\ 
& associated with a simple \\ 
& fuzzy set $(A,\mu_A)$. \\ 
$(\mu_{A,n})$ & a sequence of simple membership \\ 
& functions with monotonically \\
& finer-grained partitions.\\  

$\mathcal{U}_A$ & The fuzzy set embedding of $(A,m_A)$. \\
$\mathbb{P}(A)$ & The fuzzy measure (volume) \\ 
& of a fuzzy set under \\
 & some measure space $(U,\mathcal{F},\xi)$.\\ 
$\mathbb{P}_\mu (A)$ & The simple fuzzy measure (volume) \\
& of a fuzzy set embedding \\ 
& under some measure space $(U,\mathcal{F}, \xi)$. \\
$\mathcal{M}$ & Fuzzy mapping, which maps an input \\
& element into its associated \\
& Fuzzy Set Embedding.\\ 
$P$ & A probability measure \\
& defined on the space of concepts.
\\\bottomrule
\end{tabular}
\caption{Table for all the symbols used in this paper}
\label{table:a}
\end{table}

\section{Formal Statement of the Compactness Assumption}\label{as}

Here we state the assumption regarding the universe of discourse formally: 

\begin{assumption}\label{a:1}
The universe of discourse $U$ is \textit{topologically compact} and has an open cover. 
\end{assumption}

\begin{assumption}\label{a:2}
    The universe of discourse $U$ is measurable and is associated with a measure space $(U, \mathcal{F}, \xi)$, where $\mathcal{F}$ is the $\sigma$-algebra and $\xi$ its associated $\sigma$-finite measure. Moreover, $\forall A \in \mathcal{F}$, the fuzzy membership function $m_A: U\rightarrow [0,1]$ is $\xi$-measurable.
\end{assumption}

\section{Proof of Main Results}

In this part we provide the proof sketches of the main results: 

\textbf{Lemma 1}: \textit{Let $U$ be the universe of discourse and $A$ a fuzzy subset of $U$ with continuous membership function $m_A$, and let $\left(\mu_{A}^{(t)}\right)$ be a sequence of simple  membership functions of $m_A$ in Definition \ref{def:sfs}, such that $\{U_i\}_{i=1}^{n_{t}}$ is a refinement of $\{U_i\}_{i=1}^{n_{t-1}}$ ($n_t>n_{t-1}$), when $t$ goes to infinity, then $\mu_{A}^{(t)}$ converges to $m_A$ in the point-wise sense.}

\begin{proof}
The proof takes 2 steps:\\

\textit{Claim 1}: By construction, the sequence $\left(\mu_{A}^{(t)}\right)$ is a sequence of monotonic non-increasing function and it is bounded below by the fuzzy membership function $m_A$. 
\begin{proof}
    The case where $U$ is finite or countably infinite is straightforward. For the case where cardinality of $U$ is uncountable but $U$ is compact, we have the following: By definition \ref{def:sfs}, we have that for $i\in \{1,\cdots, d\}$ and $n\in \mathbb{N}$, $\forall u \in U_i$, $\mu_{A}^{(t)}(u) = \sup_{u\in U_i} m_A(u) \geq m_A(u)$. Hence on the entire domain, $\mu_{A}^{(t)}(u)\geq m_A(u)$. So $\forall t\in \mathbb{N}^+$, $m_{A}^{(t)}$ is an upperbound for $m_A$. Since $n_{t}>n_{t-1}$ indicates $\{U_i\}_{i=1}^{n_t}$ is a finer-grained partition than $\{U_i\}_{i=1}^{n_{t-1}}$, and the fact that supremum of a function over finer grained partition is not greater than supremum over coarse-grained partition (e.g., supremum is monotonic w.r.t partitions), we have a non-increasing sequence of functions.  
\end{proof}

\textit{Claim 2}: The sequence of functions $\left(\mu_{A}^{(t)}\right)$ converges to $m_A$. 
\begin{proof}
    Since the space $U$ is compact, we can define the space of function to be a compact Banach space of functions $f: U\rightarrow [0,1]$ and so $m_A$ is in the space. By the monotone convergence theorem \cite{realanalysis} and by the fact that supremum over a singleton set $\{x\}$ is simply $\sup_{u\in \{x\}}m_A(u) = m_A(x)$, we can conclude that convergence holds in the point-wise sense. 
\end{proof}
\end{proof}

\textbf{Theorem \ref{thm:monotone-convergence}}: \textit{Let $U$ be a compact universe of discourse and $(U,\mathcal{F},\xi)$ a measure space. Let $A$ be a fuzzy subset of $U$ and $m_A$ its membership function that's measurable. Moreover, let $\mu_A$ be its fuzzy set embedding membership function, then $\forall \epsilon > 0$, $\exists \delta > 0, d > 0$ such that if $d\delta = \xi(U)$ and $||U_i|| := \min_{i} \xi(U_i) < \delta$, we have: 
    $$
    0 < \mathbb{P}_\mu(A) - \mathbb{P}(A) < \epsilon.
    $$
}

\begin{proof}
By Lemma 1 and assumption that $m_A$ and hence all $\{\mu_{A}^{(t)}\}$ are $\xi$-measurable, we have a monotonic non-decreasing sequence of non-negative simple functions that converge point-wise to $m_A$. Then by the Monotone Convergence Theorem of simple functions \cite{realanalysis}, we have that $\int_U m_A d\xi = \lim_{t\rightarrow \infty} \int \mu_{A}^{(t)} d\xi$. Moreover, since $\mu_{A}^{(t)}\geq m_A, \forall n\in \mathbb{N}^+$, we have that $\forall \epsilon >0, \exists N \in \mathbb{N}$, such that $\forall t > N$, $\mathbb{P}_\mu(A) > \mathbb{P}(A) <\epsilon$. This is equivalent to say (by virtue of construction of simple functions in definition \ref{def:sfs}) that $\exists \delta >0, d>0$, such that $d\delta = \xi(U)$ and $||U_i||:=\min_i \xi (U_i) < \delta$ (or equivalently, the partition is sufficiently fine-grained), the conclusion holds.   
\end{proof}

Here we also provide a proof sketch for the Euclidean case, where the universe $U$ is mapped into a compact subspace $\mathcal{U}\subset \mathbb{R}^d$, and the fuzzy measure is defined as a Riemann integral in $\mathbb{R}^d$ (the Euclidean volume), which is often the case for the embedding space. In this case, the following formulation of the theorem holds:

\textbf{Theorem 1 (Euclidean Case)} \textit{Let $\Omega \subset \mathbb{R}^d$ be compact and let $(A, m_A)$ be a fuzzy set with membership function $m_A: \mathcal{U} \rightarrow [0,1]$, and let $\Omega = \bigcup_{i=1}^d \mathcal{U}_i$ be a partition and $(A,\mu_A)$ its associated partition-level fuzzy set. Then if $m_A$ is Riemann-integrable on $\Omega$, then $\forall \epsilon > 0$, $\exists \delta > 0, d > 0$ such that if $d\delta = \text{Vol}(\Omega)$ and $||\mathcal{U}_i|| := \min_{i} \text{Vol}(\mathcal{U}_i) < \delta$, we have that: 
    $$
    0 < \mathbb{P}_\mu(A) - \mathbb{P}(A) < \epsilon
    $$
    that is, the possibility of a partition-level fuzzy set is an upperbound for its underlying fuzzy set and it converges to the possibility of its underlying fuzzy set when the partition is sufficiently fine-grained.}

\begin{proof}
    The proof takes 3 main steps: Since $\Omega$ and $[0,1]$ are both compact, we  need to show that (a) under a rectangular partitions in $\mathbb{R}^d$, as partition granularity increases, the possibility $\mathbb{P}_\mu(A)$ monotonically decreases and (b) the possibility $\mathbb{P}_\mu(A)$ is an upper bound of the possibility  $\mathbb{P}(A)$. After these two are shown, we can simply invoke the standard Monotone Convergence Theorem for compact spaces and show that $\mathbb{P}_\mu(A)$ converges to $\mathbb{P}(A)$ \cite{realanalysis, Wilkins2016}. 

\textbf{Step 1}: \textit{Show that $\mathbb{P}_\mu(A)$ monotonically decreases under granular partition}. This is equivalent to show that the upper Darboux sum of a $d-$dimensional Riemann integral monotonically decreases as the partition get finer-grained. This result is Lemma 6.4 in \cite{Wilkins2016}. \par 
\textbf{Step 2}: \textit{Show that $\mathbb{P}_\mu(A) \geq \mathbb{P}(A)$}. This is equivalent to say that $d-$dimensional upper Darboux sum is an upperbound for its Darboux-Riemann integral. This result is Lemma 6.6 in \cite{Wilkins2016}. \par
\textbf{Step 3}: \textit{Show that $\mathbb{P}_\mu(A)$ converges to $\mathbb{P}(A)$}. By step 1 and step 2, we can construct a sequence of monotonically decreasing upper Darboux sums. By the Riemann-integrability of the function $m_A$ and the compactness of the set $\Omega, [0,1]$, we can conclude that by Monotone Convergence Theorem \cite{realanalysis}, this conclusion holds.
\end{proof}

\textbf{Corollary \ref{cor:4.8}} \textit{Following the condition in definition \ref{def:fuzzy-measure}, if in addition $m_A$ is Lipschitz-continuous, then the convergence rate in \ref{thm:monotone-convergence} is $O(1/n)$. If $m_A$ instead has bounded variation, then the convergence rate is also $O(1/n)$, where $n$ is the number of partitions.}
\begin{proof}
    In our case we defined the simple fuzzy set membership function as: 
\begin{align*}
    f_n(x) = \sum_{i=1}^n \sup f(x) \mathbf{1}(x\in U_i)
\end{align*}
where $(f_n)_{n\in \mathcal{N}}, f$ are measurable functions on the space $(U,\mathcal{F},\mu)$ and $\mathbf{1}$ is the indicator function, and we have $U = \bigcup_{i=1}^n U_i$ a partition of the universe. To derive a bound for convergence rate, we need to evaluate $\int |f_n - f|d\mu = \int (f_n - f) d\mu$, since $f_n \geq f$. and we have the following result:

\begin{align*}
    \int (f_n - f) d\mu &= \sum_{i=1}^n \sup f(x) \mathbf{1}(x\in U_i) - \sum_{i=1}^n \int_{U_i} fd\mu \\
    &= \sum_{i=1}^n \int_{U_i} \big(\sup_{x\in U_i} f(x) - f(x)\big) d\mu 
\end{align*}

Now suppose that  $f$ is Lipschitz continuous with constant $L$ (a much stronger assumption), then we have that 
\begin{align*}
    \forall x \in U_i, |\sup_{x\in U_i} f(x) - f(x)| \leq L \mu(U_i) 
\end{align*}
Then we have 
\begin{align*} \label{eq:4}
    \begin{split}
    \int (f_n - f) d\mu &= \sum_{i=1}^n \int_{U_i} \big(\sup_{x\in U_i} f(x) - f(x)\big) d\mu  \\
    &\leq \sum_{i=1}^n \int_{U_i} L\mu(U_i) d\mu  = \sum_{i=1}^n L\mu(U_i)^2 
    \end{split}
\end{align*}
without loss of generality, let $\{U_i\}$ be an even partition and let $\mu(U) = 1$, then we have that $\mu(U_i) \propto O(\frac{1}{n})$ and so $\mu(U_i)^2 \propto O(1/n^2)$. Hence Equation $\ref{eq:4}$ decays with $O(1/n)$ rate.

Suppose that $f$ has bounded total variation $V(f) < \infty$, then we have that: 
\begin{align*}
    \begin{split}
         \int (f_n - f d)\mu &= \sum_{i=1}^n \int_{U_i} \big(\sup_{x\in U_i} f(x) - f(x)\big) d\mu \\
         &\leq \sum_{i=1}^n \frac{V(f)}{n} \mu(U_i)
    \end{split}
\end{align*}
Again, we have the conclusion that the error decays with $O(1/n)$.

\end{proof}

\section{Details on Experiments}

\subsection{Baselines}

For our taxonomy expansion task, we compare with existing methods that use vector embeddings or geometric embeddings. For vector embedding methods, we include also models that use advanced structures of taxonomy data. To summarize, the baselines we compare with are: 
\begin{itemize}[leftmargin=*]
    \item \textbf{TAXI}\cite{panchenko-etal-2016-taxi}: This is a vector-based embedding model that relies heavily on hypernym and hyponym relations between entities.
    \item \textbf{HypeNet}\cite{shwartz-etal-2016-improving}: This is a vector-based embedding model that leverages dependency paths between entity pairs. 
    \item \textbf{Bert+MLP}\cite{Yu_2020}: This is a vector-based embedding method that uses Bert \cite{devlin2019bert} to generate entity embeddings. Bert used in this model and in our own model is fine-tuned with a smaller learning.
    \item \textbf{TaxoExpan}\cite{Shen_2020}: This is a vector-based embedding method leverages local ego-graphs to model pair dependencies. It uses graph neural networks (GNN). 
    \item \textbf{STEAM}\cite{Yu_2020}: This is a vector-based embedding method that samples dependency paths from taxonomy for better structured entity embedding. 
    \item \textbf{BoxTaxo}\cite{song-boxtaxo}: This is a geometric embedding method that uses box embedding for entities in taxonomy. It is able to capture asymmetric relationships between entities. 
\end{itemize}

\subsection{Evaluation Metrics}

For evaluating the taxonomy expansion task, we follow \cite{song-boxtaxo}. For the i-th query, denote $a_i$ the true anchor and $\hat{a}_i$ the top-1 predicted anchor and let $N$ be the total number of test samples, then the three metrics we use are the following:
\begin{itemize}[leftmargin=*]
    \item \textbf{Accuracy (ACC)}: evaluates the prediction's overall correctness
    $$
    ACC = \frac{1}{N} \sum_{i=1}^N \mathbb{I}(\hat{a}_i = a_i)
    $$
    \item \textbf{Mean Reciprocal Rank (MRR)}: evaluate the rank of the correct prediction in all predictions
    $$
    MRR = \frac{1}{N}\sum_{i=1}^N \frac{1}{rank(a_i)}
    $$
    \item \textbf{Wu \& Palmer similarity (Wu\& P)}~\cite{wu-palmer-1994-verb}: measures the semantic similarity between concepts in a taxonomy
    $$
    Wu\&P = \frac{1}{N}\sum_{i=1}^N \frac{2\times \text{depth(LCA($\hat{a}_i, a_i$))}}{\text{depth}(\hat{a}_i) + \text{depth}(a_i)}
    $$
    where LCA is the least common ancestor of two inputs and depth is the depth in the taxonomy tree.
\end{itemize}

\subsection{Implementation Detail for Base Model}
 In figure \ref{fig:lambda} and \ref{fig:margin}, the base model Fuzzy Set Embedding (FUSE) is the configuration of model with un-normalized global weights corresponding to volumes of each partition. In later ablation studies, we examine the impact of normalization on the volumes, corresponding to a choice of different measure. The number of partition of FUSE on the science dataset is $500$, while the number of partition used on the environment dataset is $350$. For training stability, we also normalize the fuzzy set embedding by their Euclidean norm before multiplication with volume weights. In addition, to make a fair comparison against baselines, we use the same optimization setup as in \cite{song-boxtaxo} and provide a version of FUSE (FUSE ($\lambda=1.0$)) with equal weights on intersection and asymmetry loss.



\section{Scope and Limitation}

This work is the first attempt to use fuzzy set to model concepts in taxonomy expansion. To examine only the effectiveness of fuzzy set representation, we only use simple neural architectures and use only the child-parent pairs.  The full capacity of FUSE should be further examined using datasets that contain First Order Logic (FOL) statements, since by construction, fuzzy sets should satisfy all the fuzzy logic axioms \cite{chen2022fuzzy}. This suggests future directions to expand taxonomy datasets with more complicated queries, and to handle more graph-structured data in social analysis and text mining \cite{ren2020beta, ren2020query2box,chen2022fuzzy,zhu2022neuralsymbolic,ju2023comprehensive}. Moreover, we can explore more explicit form of fuzzy membership function, such as a mixture of Gumbel boxes, to make the learning more concrete. 

\end{document}